\documentclass[letterpaper, 10 pt, conference]{ieeeconf}  

\usepackage{graphicx}
\usepackage{amsmath}
\usepackage{hyperref}
\IEEEoverridecommandlockouts                              

\title{\LARGE \bf
One-Shot Object Localization Using Learnt Visual Cues via Siamese Networks
}

\author{Sagar Gubbi Venkatesh$^{1}$ and Bharadwaj Amrutur$^{1, 2}$
\thanks{*This work was supported by Yaskawa Electric Corporation}
\thanks{$^{1}$Dept. of Electrical Communication Engineering, Indian Institute of Science, Bangalore 560012, India {\tt\small sagar@iisc.ac.in}}%
\thanks{$^{2}$Robert Bosch Center for Cyber-Physical Systems, Indian Institute of Science, Bangalore 560012, India {\tt\small amrutur@iisc.ac.in}}%
}

\begin{document}

\maketitle
\thispagestyle{empty}
\pagestyle{empty}

\begin{abstract}

A robot that can operate in novel and unstructured environments must be capable of recognizing new, previously unseen, objects. In this work, a visual cue is used to specify a novel object of interest which must be localized in new environments. An end-to-end neural network equipped with a Siamese network is used to learn the cue, infer the object of interest, and then to localize it in new environments. We show that a simulated robot can pick-and-place novel objects pointed to by a laser pointer. We also evaluate the performance of the proposed approach on a dataset derived from the Omniglot handwritten character dataset and on a small dataset of toys.

\end{abstract}

\section{INTRODUCTION}

To build generalist robots that can operate on novel objects that it has never been exposed to in the past, we must create vision systems that can recognize new objects from a few examples. One way to specify novel objects of interest is to use a computer interface such as touchscreen or keyboard and mouse to produce a cropped image that contains the object of interest. However, it maybe more natural or easier to specify the object of interest using visual cues like pointing at it with our hand or with a laser pointer. We consider the problem of learning to infer novel objects of interest that are being pointed at with a visual cue and then localizing this object so that the robot can operate with this novel object.

One possible approach to building such a vision system is to build a visual cue detector that can find the visual cue in an image and then to crop the object from the image using the location of the cue. This cropped image can then be used by the robot to look for and manipluate the object of interest in any environment. The problem with such a pipelined approach is that it doesn't generalize easily to new visual cues. Furthermore, the errors in successive stages add up leading to poor overall performance. Errors in localizing the visual cue could result in the object of interest not being centered in the cropped image, which potentially leads to poor performance in the next stage, whereas in an end-to-end network the latter layers naturally learn to compensate for variations in the earlier layers. This problem has been observed in a wide range of tasks from scene text detection \cite{synthetictext} to speech recognition \cite{asr} where end-to-end neural networks have outperformed earlier pipelined approaches. Thus, we seek to build an end-to-end neural network that takes in the image where the object of interest is highlighted with a visual cue and the image where the novel object must be localized and directly outputs the desired location.

\begin{figure}[!t]
    \centering
    \includegraphics[width=0.85\linewidth]{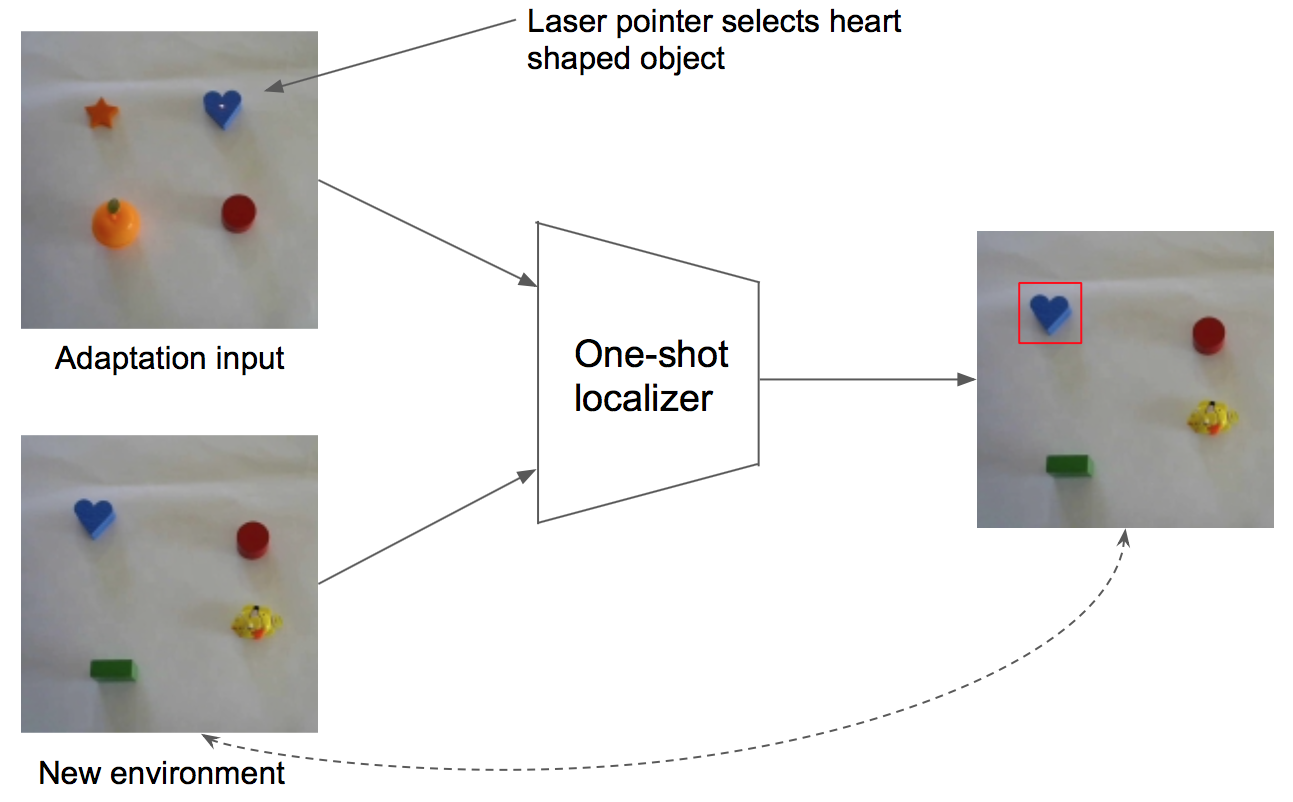}
    \caption{One-shot localization of novel objects specified by a visual cue. The laser pointer in the adaptation input specifies that the blue, heart-shaped object is of interest. The one-shot localizer then finds that object in new environments such as the one depicted in the lower image.}
    \label{introfig}
\end{figure}

Another reason to have the object localization performed with an end-to-end differentiable network is that it can then be used as part of a larger imitation learning system. Imitation learning aims to learn to control robots by imitating a human expert. Training data is gathered by recording the activity of a teleoperated robot controlled by an expert \cite{vr}. The camera feed and the current state of the robot is given to a neural network which predicts commands for the actuator.  The network architecture in this approach may be broken into two components (a) vision layers, and (b) control layers. The vision layers are a few convolutional layers followed by a spatial softmax layer that outputs 2D locations of points of interest in the image. The control layers are a few fully connected layers that consume the localization information and produce the actuation signal. In this work, we do not use a neural network for actuation. The neural network only localizes the object of interest, and the pick-and-place operation follows a pre-planned path.

In order to detect novel objects, there are broadly two approaches based on deep learning: Meta learning \cite{maml} and Siamese networks \cite{siamese}. Both approaches train on a few instances of a large number of objects with the intent of generalizing to new, unseen objects during inference. Meta-learning aims to discover initial weights for the network which when fine tuned with a few steps of gradient descent using the novel object as input results in a network that can detect the new object. Siamese networks uses twin convolutional networks with shared weights to directly learn feature representations that can discriminate between similar and dissimilar objects. We propose using a Siamese network with attention \cite{googlestreetview} to infer the new object of interest and then to find that object in a new scene (Fig.~\ref{introfig}). Note that this is a weakly supervised learning problem since the ground truth location of the visual cue is not available and only the location of the object of interest in the new environment is provided during training (Fig.~\ref{introfig}), so direct application of approaches like template matching is not possible.

The primary contribution of this paper is a system where users can specify to a robot an object of interest by using a laser pointer, which the robot can then manipulate. We evaluate our approach by having a simulated robot  pick-and-pace novel objects from a synthetic dataset. We also evaluate the performance of the one-shot localization network on a dataset derived from the Omniglot handwritten character dataset \cite{omniglot} and on a small dataset of toys.

\section{RELATED WORK}

\subsection{Meta Learning}
Meta learning, or learning to learn, takes a distribution of tasks during training and produces a quick learner that can generalize to new tasks from a few examples \cite{maml}, \cite{reptile}. Domain adaptive meta learning has been used with imitation learning to learn to imitate by observing a single demonstration performed by a human expert \cite{daml}. This differs from meta learning in that the loss function used to update the weights of the network must also be learnt because labels are unavailable for the adaptation input, and this happens through higher order derivatives. Comparing two images to determine if they are of the same object can be thought of as a distribution of tasks each corresponding to determining if the image corresponds to a particular object. When examples of a new object are given, a few gradient descent update steps adapt the neural network to look for the new object. Despite being a powerful architecture for quickly learning new tasks in a handful of iterations, the neural network is harder to train, suffers from vanishing or exloding gradients, and has a high suscpetibility to the random initialization seed as documented in \cite{howmaml}. We have been so far unsuccessful in using meta learning to localize novel objects specified by visual cues.

\subsection{Siamese Networks}
Siamese networks have been used to compare two images and determine if they have the same object \cite{siamese}, \cite{deepface}. They are twin neural networks that share weights. The feature vectors corresponding to the two images are computed by passing them through the same convolutional network. These vectors are used to perform a binary classification indicating if the two images are similar. We use the Siamese network to look for and localize a novel object of interest, but we combine it with attention so that the object of interest can be inferred from a larger scene rather than directly providing the network with the object of interest in a small image patch \cite{ox}. The attention mechanism can be thought of as matching a query to a table of key-value pairs \cite{attention}. The query is compared against all the keys through the $softmax$ operation to obtain a score for each key, and the weighted average of the corresponding values gives the value of interest. We use this to extract the feature vector of the object of interest from the feature map of a larger image in which the object is present.

A closely related method for determining similarity between images is metric learning with the triplet loss \cite{amazon}. During training, an anchor image, a positive image (similar to the anchor), and a negative image (dissimilar to the anchor) are used. The distance between the feature vectors corresponding to the positive and anchor image is minimized while that between the anchor and negative image is maximized. Although the architecture we propose is similar to the Siamese network, as we shall see in the next section, the implicit loss it uses for learning is closer to the triplet loss.

The triplet loss has been used in \cite{amazon} to sort novel objects into buckets, but the approach used is a \emph{grasp-first-then-recognize} workflow. In this work, we first localize the object of interest and then pick it up.

\section{SIMILARITY LEARNING FOR NOVEL OBJECT LOCALIZATION}

\begin{figure}[!t]
    \centering
    \vspace*{0.1cm}
    \includegraphics[width=0.85\linewidth]{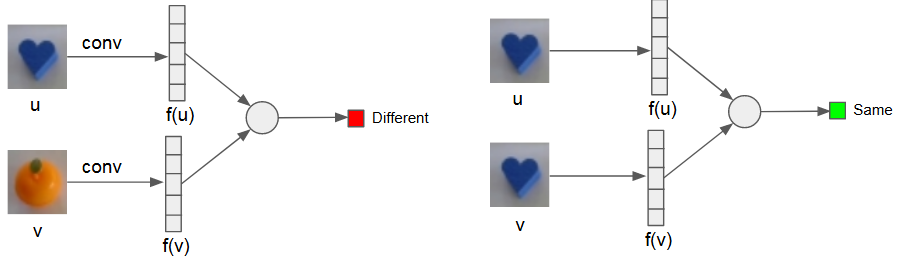}
    \caption{Illustration of a Siamese network comparing two images}
    \label{siamese_small}
\end{figure}

\begin{figure}[!t]
    \centering
    \includegraphics[width=0.85\linewidth]{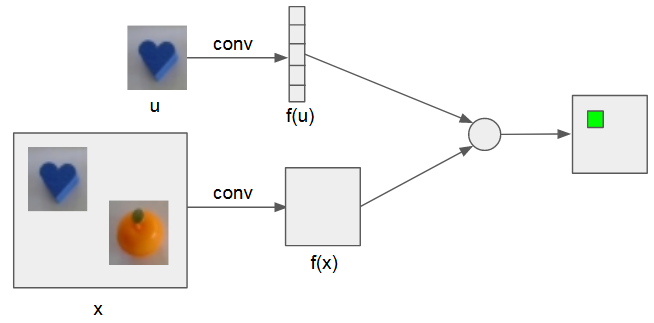}
    \caption{Illustration of localizing an object with a Siamese network}
    \label{siamese_localize}
\end{figure}

\begin{figure}[!t]
    \centering
    \includegraphics[width=0.85\linewidth]{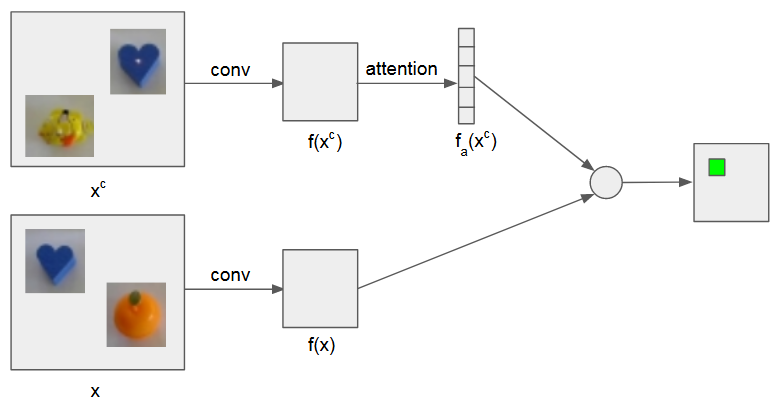}
    \caption{Illustration of using attention to extract the feature vector corresponding to the heart in the top image that is marked by a laser pointer}
    \label{siamese_attn}
\end{figure}

\begin{figure*}[!t]
    \centering
    \vspace*{0.15cm}
    \includegraphics[width=0.95\linewidth]{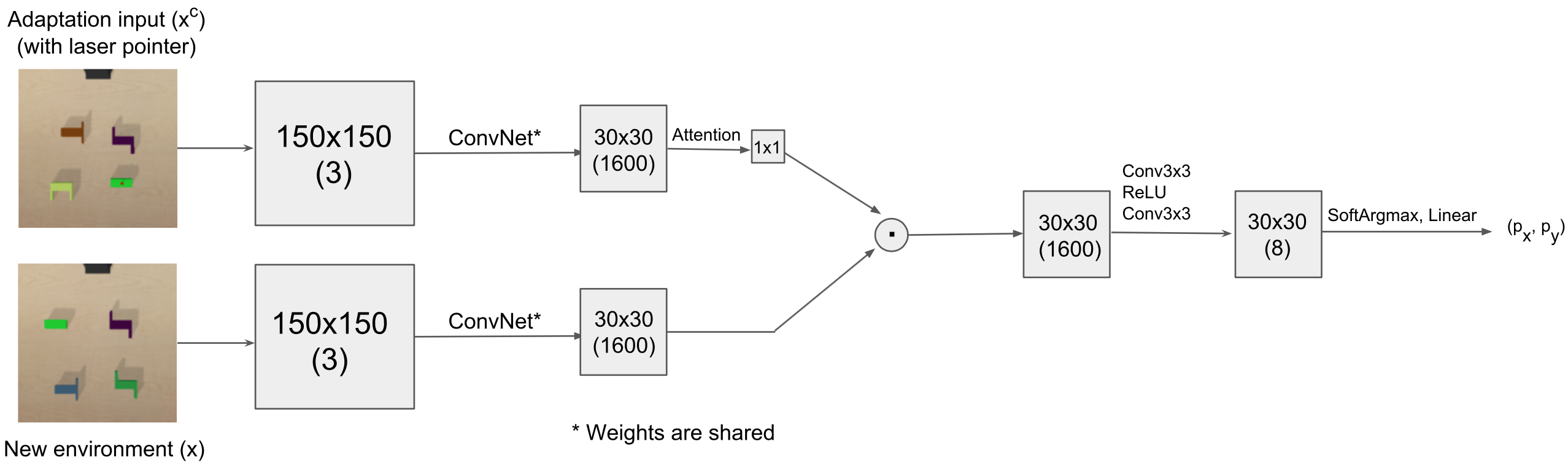}
    \caption{Illustration of the network architecture for one-shot localization. The first few convolutional layers share weights and form the Siamese network. The twins are joined with element-wise multiplication. Spatial softmax is used to extract the co-ordinates of the novel object of interest in the new environment.}
    \label{model_arch}
\end{figure*}

An image $x^c_o$ containing the object $o$ highlighted by a visual cue is drawn from the distribution $\mathcal{D}^{c}_o$. i.e., $x^c_o \sim \mathcal{D}^{c}_o$. Likewise, $(x_o, p_o) \sim \mathcal{D}_o$ where $x_o$ is an image containing the object $o$ (without any visual cue) at position $p_o$. We now have a supervised learning problem with loss function $\mathcal{L}(\theta, \mathcal{D}^c_o, \mathcal{D}_o)$ where $\theta$ denotes the model parameters.

To begin with, consider the problem of classifying if two images are similar or dissimilar\cite{siamese} (Fig.~\ref{siamese_small}). If the two images are $u$ and $v$, we would like to learn a function $\phi(u, v)$ that scores the similarity between images $u$ and $v$. One way of contructing $\phi$ is by using a convolutional network $f$. The same convolutional network $f$ is applied to both the images $u$ and $v$ to obtain feature vectors $f(u)$ and $f(v)$ respectively. The distance between these two vectors is a measure of similarity.

\begin{equation}
    \phi(u, v) = g(f(u), f(v))
    \label{small_similarity_eqn}
\end{equation}

The distance function $g$ in Eqn.~(\ref{small_similarity_eqn}) can be a simple function like the $L_2$ norm or a more complex learnable function such as a neural network.

We now turn to the problem of localizing a novel object in an image \cite{ox}. Here, the exemplar image $u$ contains the object of interest, and we would like to find the location of the object of interest in a larger image $x$ (Fig.~\ref{siamese_localize}). A natural way of addressing this problem is to have a sliding window over the larger image $x$ and to compare each window with $u$. The location of $u$ inside the large image $x$ is found when we get a matching window. If the convolutional network $f$ is fully convolutional, i.e., the network does not use padding (all the convolutions are ``valid" convolutions), then the larger image $x$ can be passed through the same convolutional network $f$ to obtain a feature map $f(x)$, where each ``pixel" in the resulting feature map will be identical to the feature vector that would have been obtained from the corresponding window in the input image. Subsequently, the feature vector $f(u)$ is compared with each pixel of the feature map $f(x)$ to determine where the object of interest lies in $x$. The similarity score for each window is,

\begin{equation}
    \phi_{i, j}(u, x) = g\Big(f(u), f_{i, j}(x)\Big)
\end{equation}

Finally, we turn to the problem where the object of interest must also be inferred from a large image (Fig.~\ref{siamese_attn}). We would like to infer which object in the image $x^c$ is being highlighted by a visual cue. We use attention to infer the highlighted object \cite{googlestreetview}. The resulting spatial attention map is used to obtain a weighted average of the feature map $f(x^c)$ that gives $f_a(x^c)$, the feature vector of interest.

We will explain procedure to obtain the feature vector of interest $f_a(x^c)$. The attention score corresponding to each pixel in the feature map $f(x^c)$ is
\begin{equation}
    a_{i, j} = h\Big(f_{i, j}(x^c)\Big)
\end{equation}
where $h$ is a small neural network comprising of Conv1x1s (bottleneck layers) and ReLUs and whose output has one channel. These attention scores are normalized to the range $(0, 1)$ with the spatial $softmax$ function.

\begin{equation}
    \alpha_{i, j} = softmax_{i, j} \big( a_{i, j} \big)
    \label{attn_map_eqn}
\end{equation}

The weighted average of the feature map computed using the normalized attention scores $\alpha_{i, j}$ gives the feature vector of the object being highlighted.
\begin{equation}
    f_a(x^c) = \sum_{i, j} \alpha_{i, j} f_{i, j} (x^c)
    \label{attn_siamese_eqn}
\end{equation}

Once the feature vector corresponding to the object of interest $f_a(x^c)$ has been obtained, localization of the object in $x$ maybe performed as described earlier.

\section{NETWORK ARCHITECTURE}

\begin{figure}[!t]
    \centering
    \includegraphics[width=0.95\linewidth]{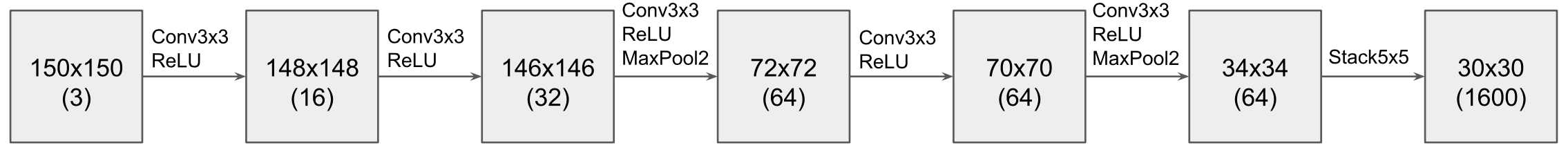}
    \caption{Fully convolutional Siamese network used in Fig.~\ref{model_arch}. Only one of the twins is depicted. The last stack layer is illustrated in Fig.~\ref{stack3x3}.}
    \label{siamese_layers}
\end{figure}

\begin{figure}[!t]
    \centering
    \vspace*{0.15cm}
    \includegraphics[width=0.75\linewidth]{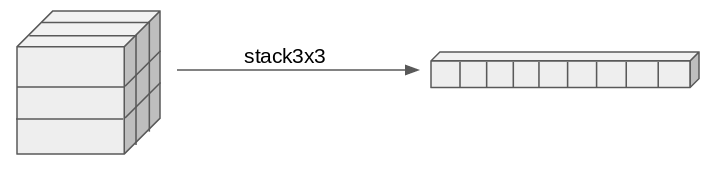}
    \caption{Illustration of the $stack3\text{x}3$ operation. It may be thought of as a convolution operation with binary weights (0 or 1) such that each output pixel is a concatenation of neighbouring pixels in the input.}
    \label{stack3x3}
\end{figure}

\begin{figure}[!t]
    \centering
    \includegraphics[width=0.95\linewidth]{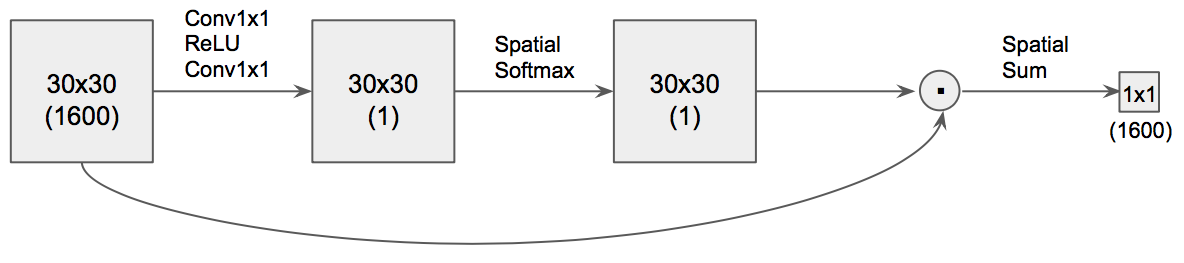}
    \caption{Spatial attention layers used to extract the feature vector corresponding to the object of interest.}
    \label{attn_layers}
\end{figure}

The proposed neural architecture for one-shot object localization is in Fig.~\ref{model_arch}. Two images are given as input to the neural network. The adaptation input $x^c$ (upper image in Fig.~\ref{model_arch}) contains the object of interest and several distracting objects. The desired object is highlighted by a visual cue. This object must be localized in target environment $x$ (lower image in Fig.~\ref{model_arch}).

The adaptation input $x^c$ is passed through the fully convolutional Siamese network (Fig.~\ref{siamese_layers}) to obtain the feature map $f(x^c)$. The spatial attention mechanism described in the previous section and detailed in Fig.~\ref{attn_layers} attends to the visual cue, and the feature vector $f_a(x^c)$ corresponding to the object of interest is extracted from the feature map. The image $x$ is passed through the same Siamese network to obtain the feature map $f(x)$. The feature vector corresponding to the object of interest $f_a(x^c)$ is combined with the feature map $f(x)$ by element-wise multiplication. The combined feature map is then passed through a few bottleneck layers $f^b$ to obtain similarity score maps. The $k$th score map is

\begin{equation}
    \phi^k_{i, j}(x^c, x) = f^b\Big(f_a(x^c) \odot f_{i, j}(x)\Big).
    \label{score_maps_eqn}
\end{equation}

We now consider how the similarity score maps $\phi^k$ may be used to localize the object of interest via a differentiable function.  If we assume that one and only one object of interest is in the image $x$, then the spatial $softargmax$ function may be used to extract the position of the object from the score maps \cite{softargmax}.


\begin{equation}
    \alpha^k_{i, j} = softmax_{i, j} \big(\phi^k_{i, j}\big)
    \label{norm_score_maps_eqn}
\end{equation}

\begin{equation}
    p^k_x = \sum_{i, j} \alpha^k_{i, j} i
    \label{softargmax_x_eqn}
\end{equation}
\begin{equation}
    p^k_y = \sum_{i, j} \alpha^k_{i, j} j
    \label{softargmax_y_eqn}
\end{equation}

Equations (\ref{softargmax_x_eqn}) and (\ref{softargmax_y_eqn}) describe the $softargmax$ operation which is merely a soft and differentiable version of the $argmax$ operation. Note that in intuitive terms, the $softargmax$ operation is asking the question ``Which window in the image is most similar to the object of interest?" and not the question ``Is this window similar to the object of interest?". This formulation is thus more similar to the triplet loss than a Siamese network based binary classifier. The distinction is relevant because the former makes the assumption that one and only one object of interest is present in the image.

The combined vector $p$ containing 2D points of interest is passed through a linear layer to obtain the location of the object of interest $(p_x, p_y)$.  Note that no supervision about the visual cue or it's location is provided during training. The inputs are the adaptation input $x^c$ containing the visual cue and the new evironment $x$, and the label is the location of the object of interest in $x$. Mean squared error is the loss function used for training.

\section{EXPERIMENTS}

We conducted experiments to evaluate the performance of the Siamese network architecture for one-shot object localization. In all the experiments, the network in Fig.~\ref{model_arch} was trained using the Adam optimizer \cite{adam} with learning rate \mbox{1e-4}.

\begin{figure}[!t]
    \centering
    \vspace*{0.15cm}
    \includegraphics[width=0.95\linewidth]{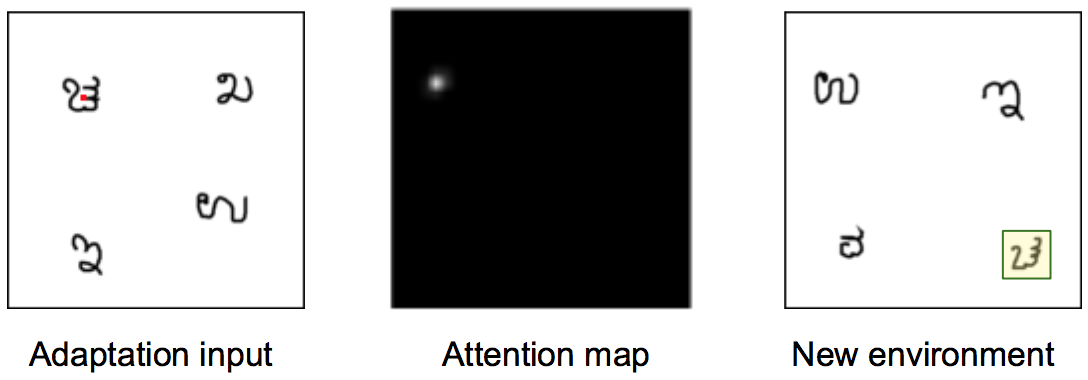}
    \caption{One-shot localization of a novel character in the Omniglot dataset. The character highlighted by the laser pointer in the adaptation input is correctly localized in the new environment. The attention map ($\alpha$ in Eqn.~(\ref{attn_map_eqn})) indicates the location of the novel character of interest in the adaptation input.}
    \label{omniglot}
\end{figure}

\begin{figure}[!t]
    \centering
    \includegraphics[width=0.95\linewidth]{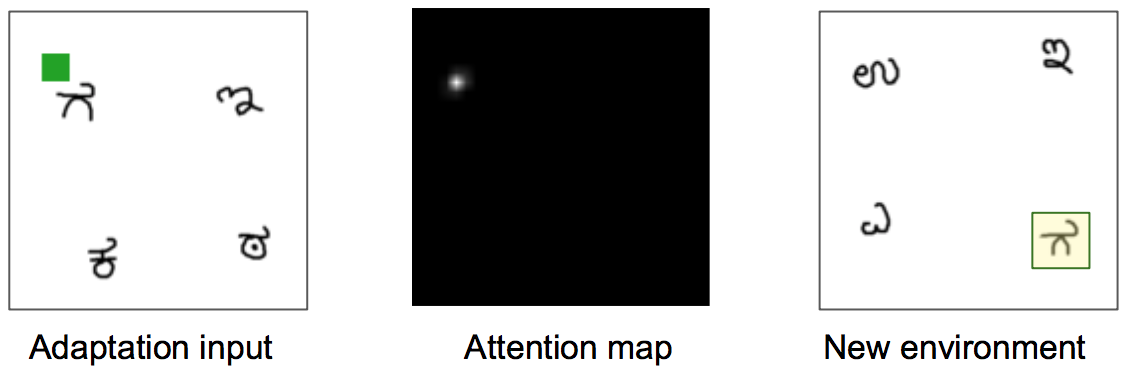}
    \caption{One-shot localization of a novel character in the Omniglot dataset. The character highlighted by the green marker in the adaptation input is correctly localized in the new environment. The attention map indicates the location of the novel character of interest in the adaptation input.}
    \label{omniglot_alt}
\end{figure}

\subsection{Localizing novel characters from the Omniglot dataset}

The Omniglot dataset \cite{omniglot} is a collection of handwritten characters of 50 alphabets from languages across the world. Each alphabet has a minimum of 15 letters to over 40 letters. There are exactly 20 handwritten instances of each letter across all the alphabets. The 50 alphabets are split into a \emph{background} set of 40 alphabets and an \emph{evaluation} set of 10 alphabets. The characters in the \emph{background} set are used for training and validation, whereas the characters from the \emph{evaluation} set are used for testing.

We considered the problem of localizing characters from the Omniglot dataset. A synthetic dataset is constructured for the localization experiment. To construct an adaptation input, instances of four different characters are chosen from the training set and placed at random non-overlapping locations on a blank canvas of 150$\times$150~px. One of the placed characters is highlighted with a red dot at the center to simulate a laser pointer. Note that the red marker is not always at the exact center of the character because of the slight variation in the way the characters have been written. The new environment is also built by starting from a blank image of 150$\times$150 px. A randomly chosen instance of the character that was previously marked is placed at a random location, and it's location is recorded. Three additional characters chosen at random either from one of the previously placed characters or a different character from the training set that is not in the adaptation input are placed at non-overlapping locations.

Figure~\ref{omniglot} shows a sample output from an example in the test set containing a novel character highlighted with a red marker. We see that the attention mechanism is focusing on the red marker, and the network has successfully localized a novel character in the new environment. With the width and height of the image normalized to 1, the mean squared error in localizing the marked character (i.e., $L2$ norm of the difference between the actual center of the character and the predicted position) across 256 examples constructed from characters in the test set is 0.002 (3.1\% of the width of the image). To evaluate the resilience of the network to small variations in the position of the red dot, we add random jitter of 33\% of the character size to the position of the red dot. Despite the added jitter, the mean squared error only slightly increases to 0.0024 on the test set. We also examine the localization performance with a different visual cue. With a large green marker above the character (Fig.~\ref{omniglot_alt}), the mean squared error on the test set is 0.003.

\subsection{Localizing novel objects on a table in PyBullet simulator}

\begin{figure}[!t]
    \centering
    \vspace*{0.15cm}
    \includegraphics[width=0.65\linewidth]{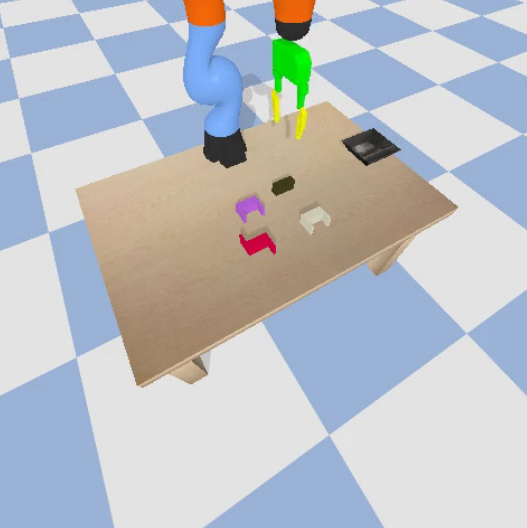}
    \caption{The Kuka robot in PyBullet. Four objects are on the table with a tray at the edge of the table where the object picked up by the robot must be placed.}
    \label{kuka}
\end{figure}

\begin{figure}[!t]
    \centering
    \vspace*{0.15cm}
    \includegraphics[width=0.65\linewidth]{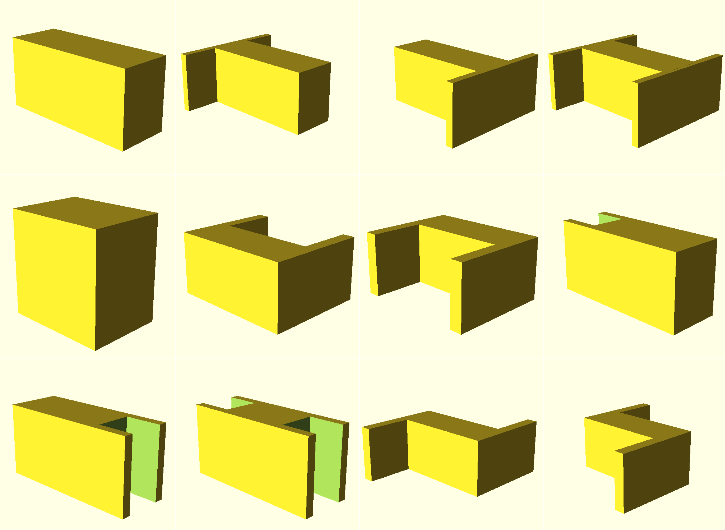}
    \caption{Shapes used for training and testing in Experiment B.}
    \label{cuboids}
\end{figure}

\begin{figure}[!t]
    \centering
    \includegraphics[width=0.95\linewidth]{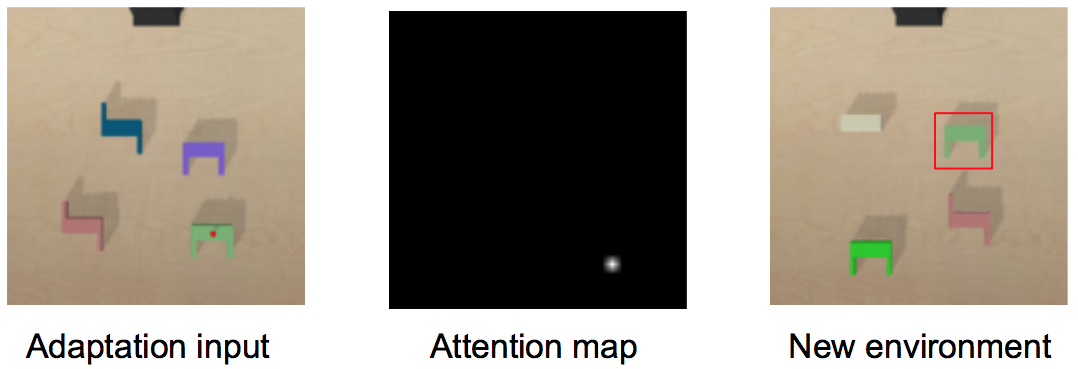}
    \caption{One-shot localization of a novel object on the table. The object highlighted by the laser pointer in the adaptation input is localized in the new environment.}
    \label{kuka_top}
\end{figure}

In order to test the proposed one-shot object localizer with a robot, we created an environment in the PyBullet physics simulator\footnote{Videos are available at \url{http://ece.iisc.ernet.in/~sagar/iros19}}. The robot used is a Kuka bot with a gripper attached so that it can pick and place objects on a table (Fig.~\ref{kuka}). We constructured 12 cuboid-like objects (Fig.~\ref{cuboids}) and used 8 of them for training and evaluation and the remaining 4 for testing. Each of the training shapes is painted with 384 random colors to obtain 3072 distinct objects for training. Likewise, the 4 shapes in the test set are each paired with 128 random colors (not intersecting with the training colors) to obtain 512 distinct test objects. Similar to the previous experiment, 4 distinct objects are spawned on the table and one of the objects is marked with a red marker (Fig.~\ref{kuka_top}). The new environment is constructured by placing the marked object at a random location along with 3 random distractor objects.

Figure~\ref{kuka_top} shows a sample output from objects in the test set. The attention layer has correctly focused on the novel object of interest in the adaptation input, and the network has successfully localized this object in the new environment. With the width and height of the image normalized to 1, the mean squared error in localizing the marked object from 256 sample test scenarios is 0.003 (3.8\% of the width of the image). The robot's operation was recorded on a 20-sample test set. In all instances, the robot successfully picked up the object and placed it in the tray at the edge of the table. To examine the peformance of this network with a smaller dataset, we reduced the number of colors in the training set to 48 and that in the test set to 12. With this truncated dataset, the network overfit, and the mean squared error in localizing objects in the test set grew to 0.010. On a test sample set of size 20, the robot successfully picked and placed 17 objects, but failed to do so in 3 cases. In one of the failure cases, the localization was off by more than 10\% which caused the gripper to collide with the object. In two other cases, the robot did not pick up the correct object.

\subsection{Localizing novel toys on a table}

\begin{figure}[!t]
    \centering
    \vspace*{0.1cm}
    \includegraphics[width=0.95\linewidth]{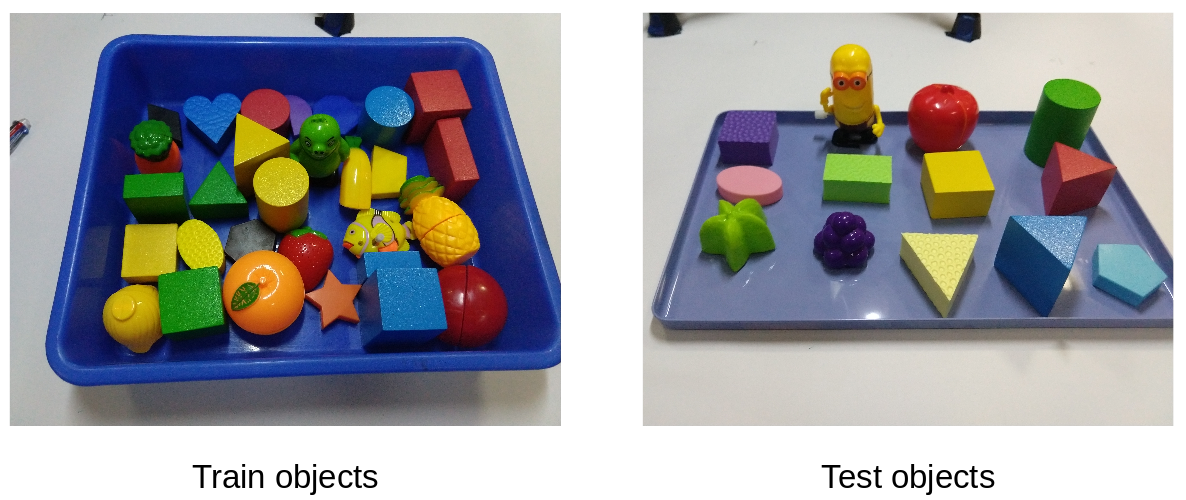}
    \caption{Objects used for training and testing in Experiment C.}
    \label{toys}
\end{figure}

\begin{figure}[!t]
    \centering
    \includegraphics[width=0.95\linewidth]{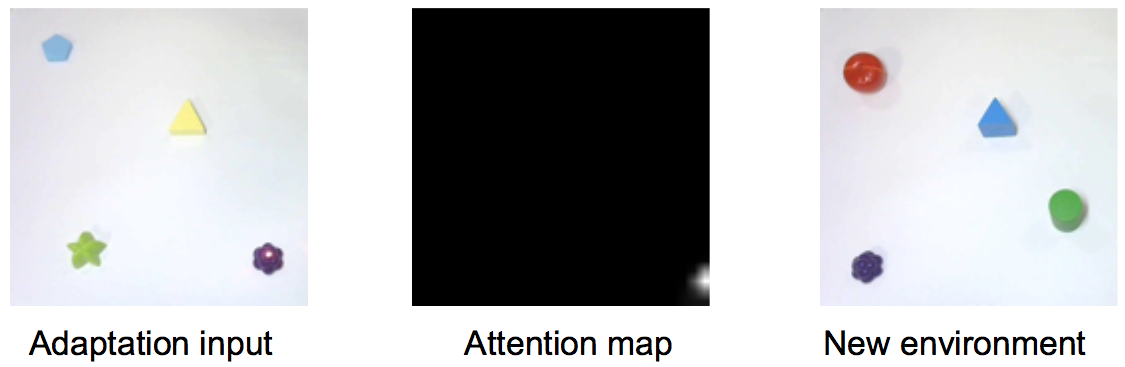}
    \caption{One-shot localization of a novel object on the table. The object highlighted by the laser pointer in the adaptation input is localized in the new environment. Note that because padding is not used convolutions, the bright spot near the corner of the attention map corresponds to the right-bottom window in the adaptation input.}
    \label{toys_top}
\end{figure}

\begin{figure}[!t]
    \centering
    \includegraphics[width=0.95\linewidth]{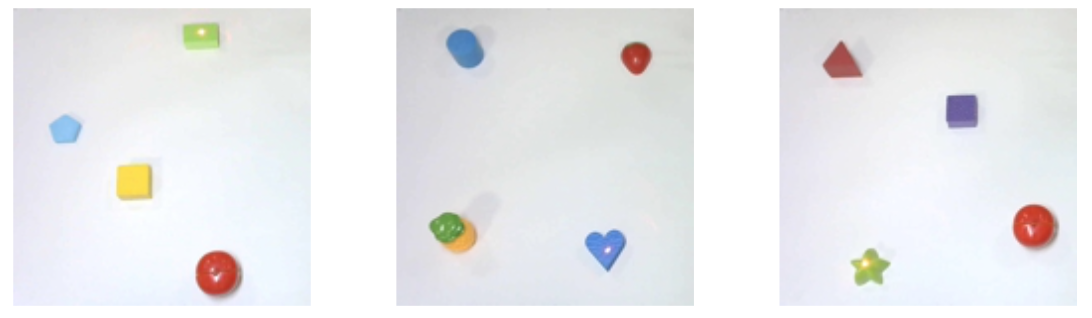}
    \caption{Specular highlights on some of the objects appear similar to the laser pointer and cause poor performance}
    \label{specular}
\end{figure}

We created a dataset of toys (Fig.~\ref{toys}) to evaluate the performance of the proposed approach on real objects. Similar to the previous experiments, four objects are placed on a table, and one of the objects is highlighted with a laser pointer (Fig.~\ref{toys_top}). The new environment consists of the highlighted object placed at a random location along with 3 random distractor objects. Figure~\ref{toys_top} shows a sample output. We see that the attention mechanism is correctly attending to the highlighted object. With the width and height of the image normalized to 1, the mean squared error in localizing the marked objects in 256 sample test scenarios is 0.03. This is considerably higher than the previous experiments. Of the 256 images, we found that the localization was successful in 216 images, but failed in the rest of the images (error $>$ 15\% of the size of the image). We found that specular highlights on the objects appeared similar to the laser pointer and had confused the attention mechanism (Fig.~\ref{specular}).  Among the successfully localized images, the mean squared error in localizing the marked objects was 0.001 (2.2\% of the width of the image).

\section{DISCUSSION}

Siamese networks are a useful tool in building robot vision systems that can adapt to novel environments. We have demonstrated that we can specify a previously unseen object of interest to a robot by using a laser pointer. This architecture can be extended to scenarios where it is desirable to specify the object of interest by pointing a finger at the object. A major limitation of this architecture is the softargmax layer which can only work in circumstances where there is one and only one object of interest in the image. Addressing this limitation is left for future work.





\section*{ACKNOWLEDGMENT}

This project was funded in part by Yaskawa Electric Corporation. Sagar Gubbi was supported by Visvesvaraya PhD fellowship, MeitY, Govt. of India. The GPU used in this work was provided by nVidia. We thank Ishan Dave and Arun Kumar for assistance with the experimental setup. We also thank Shishir N. Y. Kolathaya and Nihesh Rathod for helpful discussions.


\begin{thebibliography}{99}

\bibitem{synthetictext} Gupta, Ankush, Andrea Vedaldi, and Andrew Zisserman. "Synthetic data for text localisation in natural images." Proceedings of the IEEE Conference on Computer Vision and Pattern Recognition. 2016.
\bibitem{asr} Graves, Alex, and Navdeep Jaitly. "Towards end-to-end speech recognition with recurrent neural networks." International Conference on Machine Learning. 2014.
\bibitem{vr} Zhang, Tianhao, et al. "Deep imitation learning for complex manipulation tasks from virtual reality teleoperation." 2018 IEEE International Conference on Robotics and Automation (ICRA). IEEE, 2018.
\bibitem{maml} Finn, Chelsea, Pieter Abbeel, and Sergey Levine. "Model-agnostic meta-learning for fast adaptation of deep networks." Proceedings of the 34th International Conference on Machine Learning-Volume 70. JMLR. org, 2017.
\bibitem{siamese} Koch, Gregory, Richard Zemel, and Ruslan Salakhutdinov. ``Siamese neural networks for one-shot image recognition." In ICML Deep Learning Workshop, vol. 2. 2015.
\bibitem{googlestreetview} Wojna, Zbigniew, et al. "Attention-based extraction of structured information from street view imagery." 2017 14th IAPR International Conference on Document Analysis and Recognition (ICDAR). Vol. 1. IEEE, 2017.
\bibitem{omniglot} Lake, Brenden, et al. "One shot learning of simple visual concepts." Proceedings of the Annual Meeting of the Cognitive Science Society. Vol. 33. No. 33. 2011.
\bibitem{reptile} Nichol, Alex, and John Schulman. "Reptile: a scalable metalearning algorithm." arXiv preprint arXiv:1803.02999 (2018).
\bibitem{daml} Yu, Tianhe, et al. "One-shot imitation from observing humans via domain-adaptive meta-learning." arXiv preprint arXiv:1802.01557 (2018).
\bibitem{howmaml} Antoniou, Antreas, Harrison Edwards, and Amos Storkey. "How to train your MAML." arXiv preprint arXiv:1810.09502 (2018).
\bibitem{deepface} Taigman, Yaniv, et al. "Deepface: Closing the gap to human-level performance in face verification." Proceedings of the IEEE conference on computer vision and pattern recognition. 2014.
\bibitem{ox} Bertinetto, Luca, et al. "Fully-convolutional siamese networks for object tracking." European conference on computer vision. Springer, Cham, 2016.
\bibitem{attention} Vaswani, Ashish, et al. "Attention is all you need." Advances in Neural Information Processing Systems. 2017.
\bibitem{amazon} Zeng, Andy, et al. "Robotic pick-and-place of novel objects in clutter with multi-affordance grasping and cross-domain image matching." 2018 IEEE International Conference on Robotics and Automation (ICRA). IEEE, 2018.
\bibitem{softargmax} Levine, Sergey, et al. "End-to-end training of deep visuomotor policies." The Journal of Machine Learning Research 17.1 (2016): 1334-1373.
\bibitem{adam} Kingma, Diederik P., and Jimmy Ba. "Adam: A method for stochastic optimization." arXiv preprint arXiv:1412.6980 (2014).


\end{thebibliography}
\end{document}